\documentclass[journal]{IEEEtran}
\usepackage{graphicx}
\usepackage{subfigure}
\usepackage{array}
\usepackage{amsmath,epsfig}
\usepackage{booktabs}
\usepackage[ruled]{algorithm2e}
\usepackage{algorithmic}
\usepackage{algorithm2e}
\usepackage{multirow}
\usepackage{diagbox}
\usepackage[table]{xcolor}
\usepackage{tabularray}
\usepackage{amssymb}
\usepackage{booktabs}
\usepackage{float}
\usepackage{cite}
\usepackage[driverfallback=dvipdfm,
CJKbookmarks=true, bookmarksnumbered=true, bookmarksopen=true,
colorlinks=true,
pdfborder=001,
citecolor=blue, linkcolor=blue, anchorcolor=red, urlcolor=black
]{hyperref}
\begin{document}\sloppy
\title{Improving Adversarial Robustness via Decoupled Visual Representation Masking}

\author{Decheng~Liu,~Tao~Chen,~Chunlei~Peng,~\IEEEmembership{Member, IEEE},~Nannan~Wang,~\IEEEmembership{Senior Member, IEEE},~Ruimin~Hu,~Xinbo~Gao,~\IEEEmembership{Fellow, IEEE}

\noindent
\thanks{D. Liu, T. Chen, C. Peng and R. Hu are with the State Key Laboratory of Integrated Services Networks, School of Cyber Engineering, Xidian University, Xi’an 710071, Shaanxi, P. R. China (e-mail: dchliu@xidian.edu.cn; tchen.xidian@gmail.com; clpeng@xidian.edu.cn; rmhu@xidian.edu.cn).\\
N. Wang is with the State Key Laboratory of Integrated Services Networks, School of Telecommunications Engineering, Xidian University, Xi’an 710071, Shaanxi, P. R. China (e-mail: nnwang@xidian.edu.cn).\\
X. Gao is with the Chongqing Key Laboratory of Image Cognition, Chongqing University of Posts and Telecommunications, Chongqing 400065, P. R. China.(e-mail: gaoxb@cqupt.edu.cn).}

}


\markboth{Journal of \LaTeX\ Class Files,~Vol.~16, No.~8, June~2024}%
{Shell \MakeLowercase{\textit{et al.}}: A Sample Article Using IEEEtran.cls for IEEE Journals}


\IEEEcompsoctitleabstractindextext{%
\begin{abstract}
Deep neural networks are proven to be vulnerable to fine-designed adversarial examples, and adversarial defense algorithms draw more and more attention nowadays. 
Pre-processing based defense is a major strategy, as well as learning robust feature representation has been proven an effective way to boost generalization.
However, existing defense works lack considering different depth-level visual features in the training process.
In this paper, we first highlight two novel properties of robust features from the feature distribution perspective:
1)	\textbf{Diversity}. The robust feature of intra-class samples can maintain appropriate diversity;
2)	\textbf{Discriminability}. The robust feature of inter-class samples should ensure adequate separation.
We find that state-of-the-art defense methods aim to address both of these mentioned issues well.
It motivates us to increase intra-class variance and decrease inter-class discrepancy simultaneously in adversarial training.
Specifically, we propose a simple but effective defense based on decoupled visual representation masking.
The designed Decoupled Visual Feature Masking (DFM) block can adaptively disentangle visual discriminative features and non-visual features with diverse mask strategies, while the suitable discarding information can disrupt adversarial noise to improve robustness.
Our work provides a generic and easy-to-plugin block unit for any former adversarial training algorithm to achieve better protection integrally.
Extensive experimental results prove the proposed method can achieve superior performance compared with state-of-the-art defense approaches. 
The code is publicly available at 
\href{https://github.com/chenboluo/Adversarial-defense}{https://github.com/chenboluo/Adversarial-defense}.

\end{abstract}

}

\maketitle
\IEEEdisplaynotcompsoctitleabstractindextext
\IEEEpeerreviewmaketitle

\section{Introduction}

Deep learning has been widely applied in numerous real-world fields, such as face recognition~\cite{zhao2003face,su2023hybrid,wu2023face}, text generation~\cite{lin2023tavt,bayer2023data,zhang2019bertscore}, and autonomous driving~\cite{yurtsever2020survey,yurtsever2020survey,alibeigi2023zenseact}. 
However, researchers find that neural networks are vulnerable to adversarial examples~\cite{goodfellow2014explaining}, which are crafted maliciously based on raw inputs and can easily fool target models into making wrong predictions by adding imperceptible noise. 
Thus, it poses a significant security risk~\cite{9519486,10.1145/3548606.3560671,zhang2023apmsa} for the application of deep learning, especially in decision-critical scenarios. 

\begin{figure*}

\begin{center}
\subfigure[PGD Attack.]
{\includegraphics[width=0.4\linewidth]{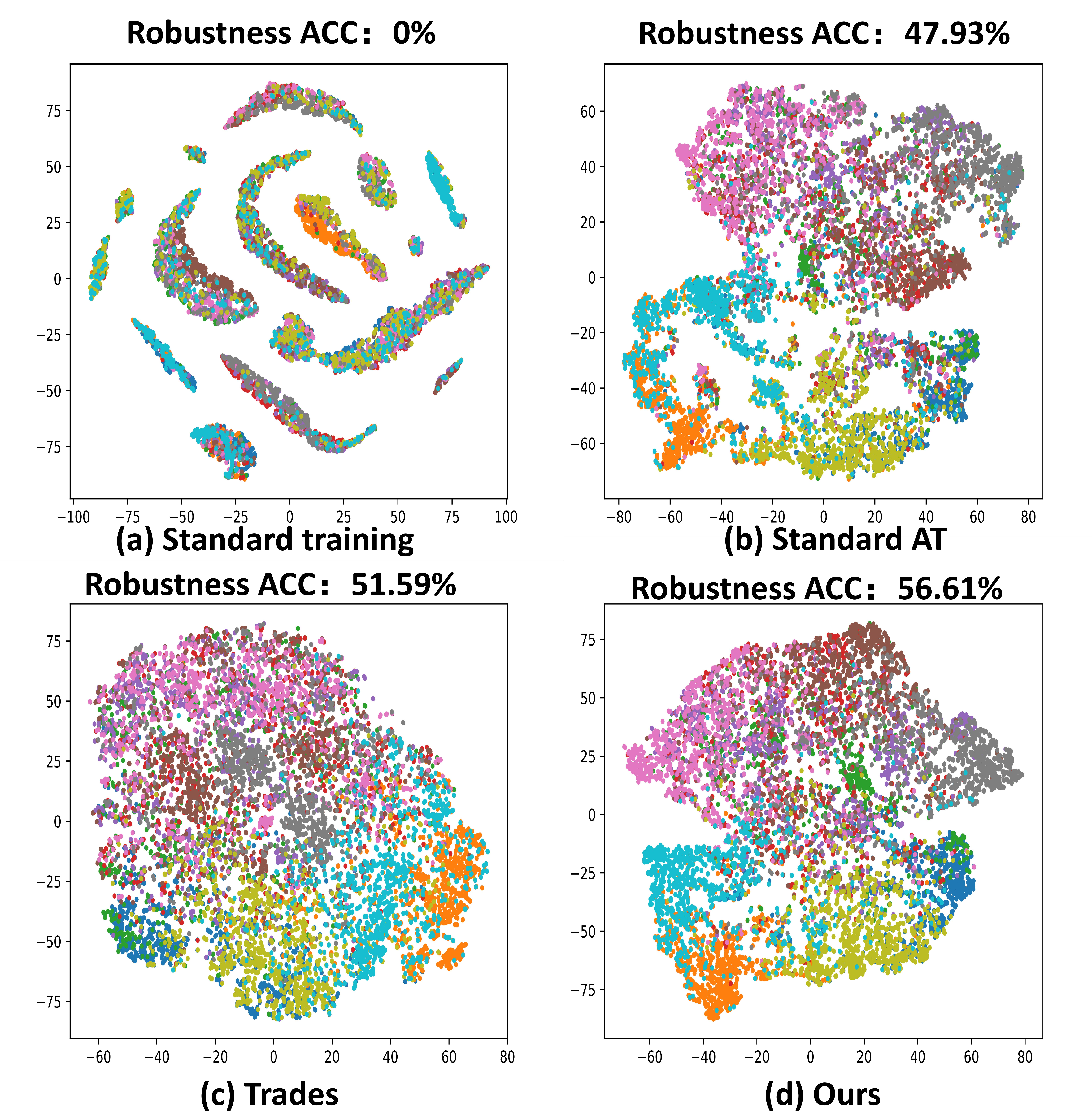}} 
\quad \quad \quad 
\subfigure[Auto Attack.]
{\includegraphics[width=0.4\linewidth]{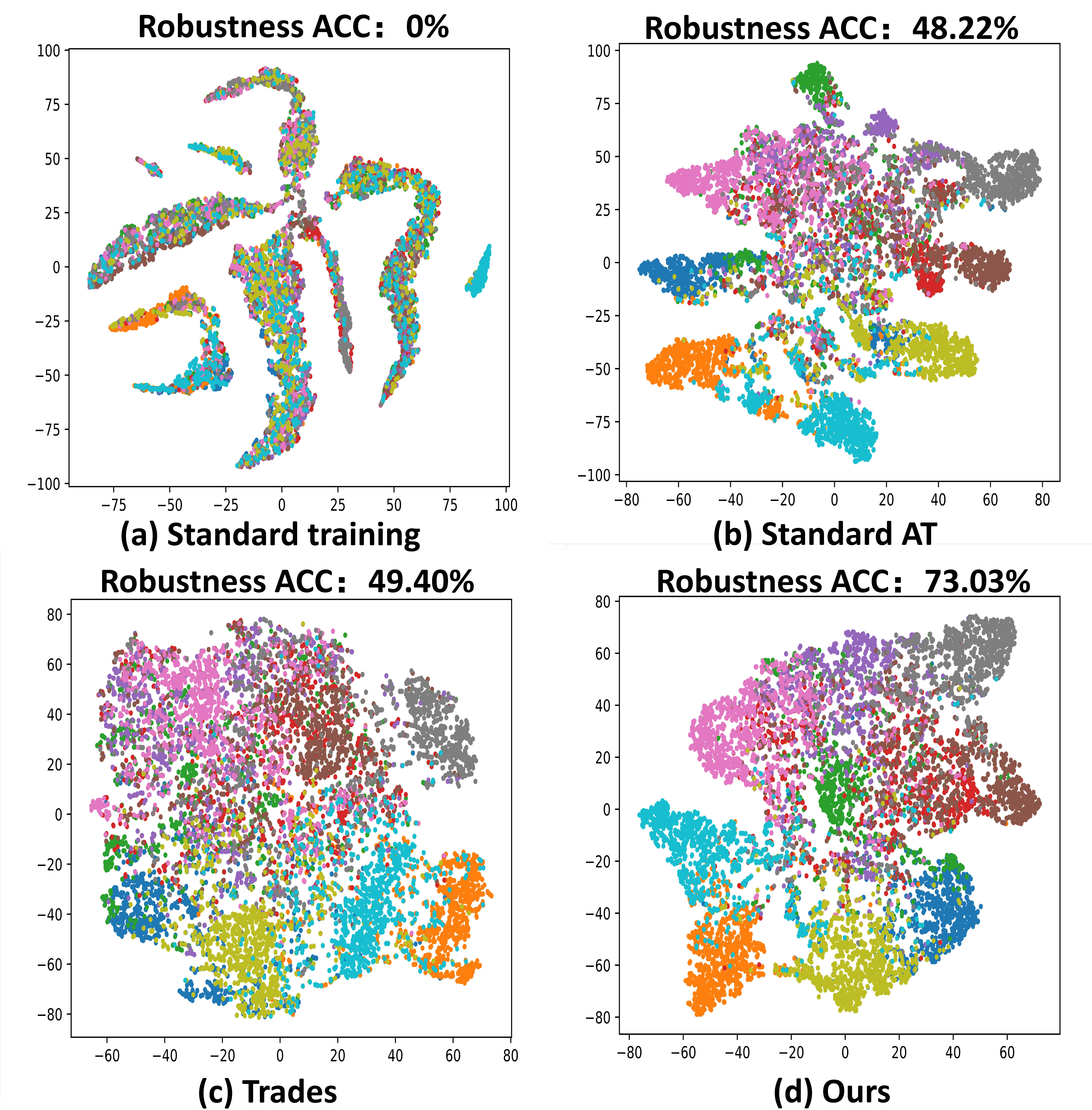} }
\caption{(a) the t-SNE visualizations of PGD adversarial example on
different methods. (b) the t-SNE visualizations of Auto Attack adversarial example on
different methods.
The robust feature of intra-class samples should maintain appropriate diversity, and the robust feature of inter-class samples should ensure adequate separation.}
\label{fig: motive}
\end{center}  
\end{figure*}

Existing researchers generally adopt the Adversarial Training (AT) strategy as a typical defense approach~\cite{zhang2019theoretically,wu2020adversarial}.
It utilized generated adversarial examples as data augmentation during model training, and the model can learn the characteristics of adversarial examples and enhance its robustness. 
However, adversarial training-based defense methods need to select adversarial examples according to possible attack scenarios, resulting in the poor generalization ability of the model when facing unseen attacks ~\cite{zhou2021towards}.
The pre-processing strategy~\cite{guesmi2021sit,raff2019barrage,xu2022mask} is another major class of adversarial defense, which has high scalability~\cite{Kim_2023_CVPR} but seems vulnerable to advanced attacks using proxy gradients~\cite{zimmermann2019comment,lee2022graddiv}.
Adversarial defense based on feature space can help effectively improve the robustness and generalization~\cite{ozfatura2021less,Kim_2023_CVPR}. 
However, most existing methods lack consideration of different depth-level features in the robust feature space.

In the paper, \emph{we first show the strong connections between characteristics of feature distribution and adversarial robustness from a novel perspective.}
The feature distribution analysis is motivated by the phenomenon that different feature distribution characteristics could result in discrepant robust performance when against adversarial examples (as shown in Fig. \ref{fig: motive}).
Compared with the standard training, the advanced defense algorithms seem to suitably decrease intra-class invariance in the perspective of feature distribution.
This is because the appropriate intra-class invariance could help improve the diversity to alleviate the overfitting.
Additionally, we also find that adequate inter-class discrepancy constraints can prevent misclassification.
The proof--conception experimental results show that \emph{the good robust feature distribution should simultaneously maintain suitable degrees of both diversity and discriminability.}

To address these issues, we propose a novel generic defense algorithm based on decoupled visual representation masking.
From the perspective of feature distribution, we aim to increase intra-class invariance and decrease inter-class discrepancy simultaneously in the adversarial training stage.
Specifically, we design a Decoupled Visual Feature Masking (DFM) block, which can adaptively disentangle visual discriminative features and non-visual features according to the diverse masking strategies (Section~\ref{Visualization1}).
Different from~\cite{Kim_2023_CVPR}, the proposed decoupled strategy doesn't need an extra training process with specific constraints.
The proposed visual masking strategy can help effectively improve the diversity of extracted robust features.
Moreover, the following non-visual feature masking and fusion strategy not only disrupts adversarial noise but also ensures the integrality of discriminative information.
Additionally, the designed DFM block unit can be easily plugged into existing adversarial training methods to improve robustness.

The main contributions of our paper can be summarized as follows:

\begin{enumerate}
    \item We provide a novel feature distribution perspective for adversarial defense and two ideal properties of good robust feature: 1) the robust feature of intra-class samples need to maintain appropriate diversity, and 2) the robust feature of inter-class samples should ensure adequate separation.
    Thus, it inspires researchers to increase intra-class variance and decrease inter-class discrepancy simultaneously for better robustness.
    
    \item We propose a newly easy-to-plugin unit to train the robust network via a Decoupled Visual Feature Masking (DFM) block. In the training stage, the DFM block can adaptively disentangle visual discriminative features and non-visual features with diverse mask strategies, and effectively disrupt adversarial noise.

    \item Experimental results on multiple public datasets illustrate the superior performance of the proposed method compared with the state-of-the-art adversarial defense algorithm. Meanwhile, the designed DFM block is generic and can be easily incorporated into many existing defense methods to boost performance. The code is publicly available at 
\href{https://github.com/chenboluo/Adversarial-defense}{https://github.com/chenboluo/Adversarial-defense}.
    
\end{enumerate}


\section{Related work}

\subsection{Adversarial Attack}
Szegedy \textit{et al.}
\cite{szegedy2013intriguing} first revealed that adversarial examples can mislead deep neural networks. Goodfellow \textit{et al.} \cite{goodfellow2014explaining} proposed the fast gradient sign method attack (FSGM), and Madry \textit{et al.} 
\cite{Madry2018Improving} proposed the projected gradient descent (PGD) attack. 
Croce \textit{et al.} \cite{croce2020reliable} overcame the failure of PGD due to suboptimal step size and objective function and proposed the auto-PGD (APGD) attack. They also combined the new attack with two complementary existing attacks, forming a parameter-free, computationally affordable, and user-independent attack ensemble (AA). 
Carlini \textit{et al.} 
\cite{carlini2017towards} customized the adversarial attack algorithm C\&W attack based on three distance metrics. 
Rony \textit{et al.} \cite{rony2019decoupling} improved the C\&W attack by decoupling the direction and norm of the adversarial perturbation. They also proposed the gradient-based attack DDN attack. 
Zimmermann \textit{et al.} 
\cite{zimmermann2019comment} proposed adjusting the adversarial attack to combine the stochastic nature of Bayesian networks to evaluate their robustness (EOTPGD) accurately. Many pre-processing methods have a large performance degradation under this attack.

\subsection{Adversarial Defense}

\textbf{Adversarial Training.} Adversarial training is a method to enhance the adversarial robustness of models by using adversarial examples. It formulates the training process as a min-max optimization problem, aiming to resist adversarial perturbations. 
Goodfellow \textit{et al.}
\cite{goodfellow2014explaining} first proposed to add adversarial examples generated by FSGM to the training phase, but this method is vulnerable to iterative attacks. To overcome this drawback, Madry \textit{et al.} \cite{Madry2018Improving} proposed an adversarial training method based on PGD, which can withstand stronger adversarial attacks.
Moreover, the training strategy also affects the effectiveness of adversarial training. For example, Zhang \textit{et al.} \cite{zhang2019theoretically} proposed a TRADES defense method based on the trade-off theory between natural accuracy and adversarial robustness. 
Wang \textit{et al.} \cite{Wang2020Improving} investigated the impact of misclassified and correctly classified examples on the final robustness of adversarial training and proposed a misclassification-aware adversarial training (MART) algorithm. Huang \textit{et al.} \cite{Huang_2023_CVPR} designed an adaptive adversarial distillation (AdaAD) method that interacts between teacher and student models. In addition, data augmentation is also an effective means to improve robustness. 
Li \textit{et al.} \cite{li2023revisiting} proposed the introduction of a biased distribution strategy on the positive gradient to maintain the dominant role of significant true class gradients during the learning process.
Kuang \textit{et al.} \cite{kuang2024defense} encouraged maintaining the consistency of the topological structure within the feature space of both natural and adversarial samples during model training.
However, these adversarial training-based methods have some limitations, such as the manually selected adversarial examples may only cover some possible attack modes, resulting in insufficient generalization ability of the model when facing unknown attacks.

\textbf{Pre-processing Defense.} Pre-processing defense is a defense method that transforms the data at the input layer to reduce the impact of adversarial perturbations. Raff \textit{et al.} 
\cite{raff2019barrage} combined many weak defense methods into a single random transformation defense method to increase the randomness and diversity of defense.  Guesmi \textit{et al.}
\cite{guesmi2021sit} proposed a random input transformation (SIT) method, which destroys the structure of adversarial perturbations by randomly rotating, cropping, scaling, etc., the input data. 
Berger \textit{et al.}
\cite{berger2022stereoscopic} proposed an attack method to evaluate the random transformation defense. They also used the new attack to train the RT defense adversarially (AdvRT), achieving a significant robustness gain. Blau \textit{et al.}
\cite{blau2023classifier} introduced a method to enhance the robustness of the network by testing time transformation (TETRA). This novel defense method leverages PAG to improve the performance of the trained robust network. 
Xiang \textit{et al.}
\cite{xiang2022patchcleanser} performed two rounds of pixel masking on the input image to counteract the effect of the adversarial patch. 
Pre-processing defense has high scalability but seems vulnerable to advanced attacks using proxy gradients~\cite{zimmermann2019comment,lee2022graddiv}.

\textbf{Defense on Feature Space.} 
In recent years, researchers have begun to explore 
feature-level pre-processing to increase adversarial robustness. Xu \textit{et al.} 
\cite{xu2019interpreting} analyzed the relationship between adversarial robustness and class activation mapping localization regions. They proposed that sensitive feature units are masked to increase the adversarial robustness.
Xie \textit{et al.} \cite{xie2019feature}  improved adversarial robustness by performing feature denoising. Yan \textit{et al.} 
\cite{yan2021cifs} introduce Channel-wise Importance-based Feature Selection to reduce the impact of non-robust features.
Ozfatura \textit{et al.} 
\cite{ozfatura2021less} selected features by observing the consistency of activation values in the penultimate layer. Kim \textit{et al.} \cite{Kim_2023_CVPR} improved adversarial robustness by separating robust features and recalibrating malicious features. However, most existing methods need more consideration of depth-level features in the robust feature space.

In contrast to these methods, we propose a new easy-to-plugin unit to disentangle visual discriminative and non-visual features adaptively according to the diverse masking strategies.
Meanwhile, we propose a non-visual feature masking and fusion strategy that not only disrupts adversarial noise but also ensures the integrality of discriminative information, which increases adversarial robustness.

\section{Proposed Method}

\begin{figure}[t]
\begin{center}
\includegraphics[width=0.8\linewidth]{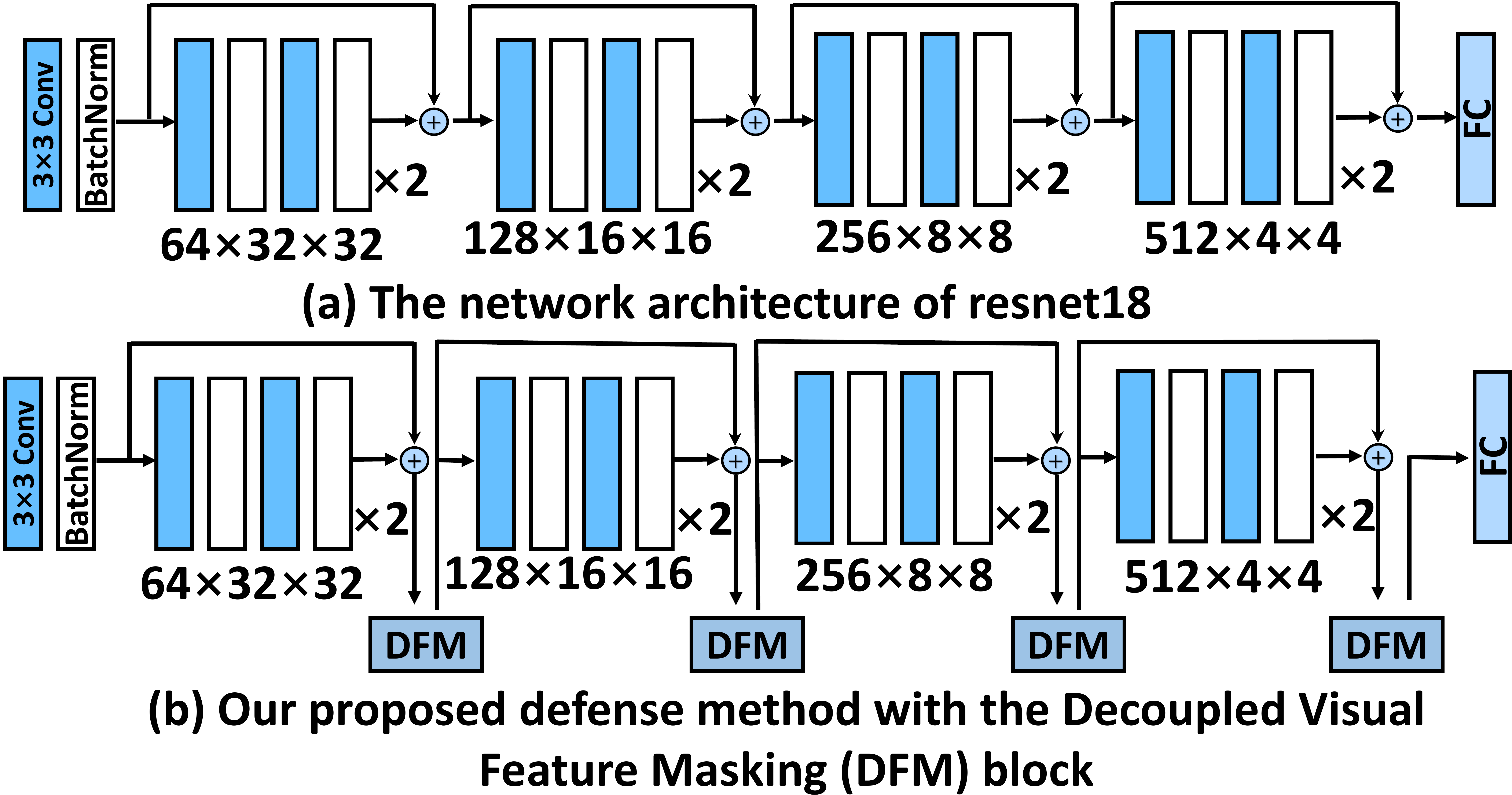} 
\caption{(a) the network architecture of ResNet18, (b) our proposed defense method with the Decoupled Visual Feature Masking block. We present the number of each residual module as well as the feature map size. The main difference between our method and ResNet18 is the addition of different Decoupled Visual Feature Masking block
 at different levels.}
\label{fig: overview1}
\end{center}  
\end{figure}

In this work, we aim to design a defense method that can make extracted features maintain suitable diversity and discriminability. Our basic intuition is: 
1). Visual masking strategy can help effectively improve the diversity of extracted robust features; 
2). Non-visual feature masking and fusion strategy not only disrupt adversarial noise but also ensures the integrality of discriminative information. 
To enhance the defense ability of varying levels of the network, we design this method as an easy-to-plugin unit and embed it between different layers of the network. 
For the convenience of illustration, we choose ResNet18 as the example to show the proposed DFM block (as shown in Fig~\ref{fig: overview1}).
The proposed DFM block can be easily extended to other backbone networks.

Given a classification dataset ${(x_i,y_i )}_{i=1}^N$, $x_i$ represents an input image of $R^{H \times W \times 3}$, where $H$ and $W$ are the height and width respectively. $y_i \in \left\{0,1,…, M\right\}$ represents the classification label, and $N$ is the total number of training images. 
$\theta$ is the parameter of target network and $\varphi$ is the total parameter of easy-to-plugin architectural unit. Our training objective is to find a robust network and corresponding easy-to-plugin architectural unit $F_{\theta, \varphi}(\cdot)$. $\delta$ is the adversarial noise.

\subsection{Decoupled Visual Feature Masking Block}
\begin{figure}[t]
\begin{center}
\includegraphics[width=0.95\linewidth]{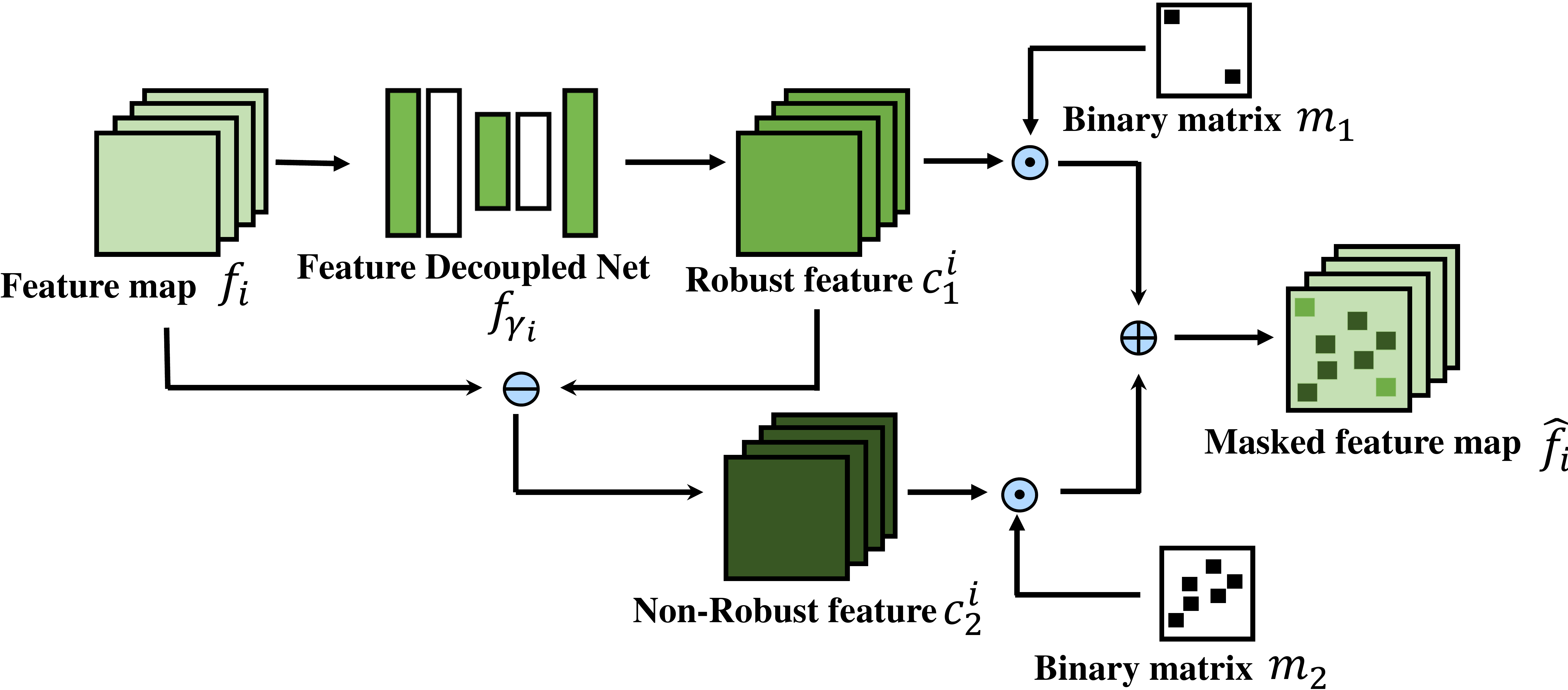}
\caption{Detailed design of the Decoupled Visual Feature Masking Block. Specifically, we decouple visual discriminative features from non-visual features by Feature Decoupled Net. Then, we mask them separately with different masking rates and combine the enhanced features.
}
\label{fig: overview2}
\end{center}  
\end{figure}

In this section, we present the Decoupled Visual Feature Masking (DFM) block as illustrated in Figure~\ref{fig: overview2}. It divides the features into two parts by using a simple three-layer network (Feature Decoupled Net) for feature decoupling: visual discriminative features and non-visual features. Specifically, the visual discriminative features are consistent with human perception, which can help the classification task.  
The consistency analysis of visual features and human perception is shown in Section~\ref{Visualization1}.
Then, we multiply the visual discriminative features and non-visual features by a random binary matrix $M_1$ and $M_2$ 
with mask ratios of $r_1$ and $r_2$ to disrupt adversarial noise and ensure the integrality of discriminative information. 
\emph{It's worth noting that separating the two features does not need a specific loss function, which only depends on the masking strategies.} 
We need to adjust the masking ratios $r_1$ and $r_2$ for visual discriminative features and non-visual features. 
For the visual feature extracted branch, we set lower mask ratios to enhance the diversity of robust features; while for the non-visual feature extracted branch, we set higher mask ratios to disrupt adversarial noise to boost robustness. 
Finally, we fuse both two disentangled features to maintain the integrality of discriminative information.
The corresponding parameter analysis is shown in Section~\ref{Parameter Analysis}.

Assuming we divide the target network into $K$ blocks. For $i_{th}$ block, we design DFM architectural unit $\varphi_i$  to separate the visual discriminative features and non-visual discriminative features.
Given an adversarial example $x_{adv} = x+\delta$ is input to the network and the feature $f_i$ of the $i_{th}$ block is extracted. 
For the feature maps of the $i_{th}$ block, we first decouple the visual feature $c_1^i$ through the Feature Decoupled Net: 
\begin{equation}
c_1^i=F_{\varphi_i} (f_i ). ~\label{eq 2}
\end{equation}
 
Then, we subtract the original feature from the visual discriminative features to get the residual non-visual discriminative features $c_2^i$ as follows:
\begin{equation}
c_2^i=f_i-F_{\varphi_i}(f_i). ~\label{eq 3}
\end{equation}

We mask the visual discriminative features $c_1^i$ and the non-visual feature $c_2^i$ respectively, by multiplying the random binary matrix $M_1$ and $M_2$. Where the mask ratio of $M_1$ is $r_1$ and the mask ratio of $M_2$ is $r_2$. 
$r_1$ and $r_2$ are essential parameters of the random mask $M_1$ and $M_2$. 
More detailed analysis is shown in Section~\ref{Parameter Analysis}.
We find that visual discriminative features will adaptively flow to branches with a lower mask ratio. 
The fused robust feature is extracted as follows:

\begin{equation}
\hat{f_i} = c_1^i \odot M_1+c_2^i \odot M_2.  ~\label{eq 4}
\end{equation}

Here $\odot$ means the Hadamard product operation.
The detailed design of the Decoupled Visual Feature Masking block can be found in Figure~\ref{fig: overview2}. At the same time, we find that different levels of feature maps have different effects on robust performance, and the more experimental analysis will be shown in section~\ref{Parameter Analysis}.

Since adversarial training can bring better adversarial robustness to the model, we first use standard adversarial training to train a target-AT (adversarial training) model as follows, where $L_{cls}$ is cross-entropy loss: 
\begin{equation}
\min_\theta   E_{(x,y)\sim D} [ \max_\delta (L_{cls} (F_\theta (x + \delta),y)) ].
\label{eq1}
\end{equation}

We further optimize the Decoupled Visual Feature Masking blocks $F_\varphi$ and the target-AT model based on the formula \ref{eq1}, when given an adversarial example $x_{adv} = x+\delta$. 
The following training stage can also be regarded as a min-max process:

\begin{equation}
\min_{\theta , \varphi}   E_{(x,y)\sim D} [\max_\delta (L_{cls} (F_{\theta , \varphi} (x + \delta),y)) ].
\end{equation}

We present a detailed settings analysis in the
following section, and the detailed algorithm of our method is shown in Algorithm~\ref{algorithm}.

\section{Experiments}
\subsection{Experimental settings}\label{Experiment setup}

\textbf{Datasets.} We select several well-known classification datasets widely used for adversarial robustness research to evaluate our method, including the CIFAR-10, CIFAR-100, and Tiny-imagenet datasets. 
CIFAR-10 dataset is one of the most widely used machine learning research datasets. It consists of 50,000 32$\times$32 training images and 10,000 test images. The CIFAR-10 dataset is divided into ten classes, each with 6,000 images. 
CIFAR-100 dataset has 100 classes. Each class has 600 32$\times$32 color images, of which 500 are used as a training set and 100 as a test set. 
Tiny-imagenet dataset has 200 classes. Each class has 500 training images, 50 validation images, and 50 test images.
MNIST dataset has 10 classes. MNIST dataset has 60000 training images, and 10000 test images.

\begin{algorithm}[!t]
	\caption{Training Details}\label{algorithm}
	\KwIn{  Target network with parameter $\theta$, the target network is divided into $K$ blocks, plugin units with parameter $\varphi$, the epoch number $E_1$ in the training phase $1$, the epoch number $E_2$ in the training phase $2$, batch size $n$, training dataset $D$,  and perturbation budget $\epsilon$. 
 }.
	\begin{algorithmic}[1]
        \WHILE{$ e < E_{1} $}
            \STATE Load mini-batch $B = \left\{x_i\right\}_{i=1}^n$ from training set $D$;
            \STATE Craft white-box adversarial examples $\hat{B}$ with perturbation budget $\epsilon$ for target network $\theta$;
            \STATE Calculate the cross entropy loss $L_{cls}$ for $\hat{B}$;
            \STATE Back-pass and update $\theta$;
        \ENDWHILE

        \WHILE{$ e < E_{2} $}
            \STATE Load mini-batch $B = \left\{x_i\right\}_{i=1}^n$ from training set $D$;
            \STATE Craft white-box adversarial examples $\hat{B}$ with perturbation budget $\epsilon$ for target network $\theta$ and plugin units $\varphi$;
            \FORALL{$i = 1$ to  $n$  (in parallel)}
                \STATE Generate random binary matrix $M_1,M_2$ with the shape of $[c_k,h_k,w_k]$;
                
                \FORALL{$k = 1$ to $K$}
                    \STATE Get feature map $f_k$ of the $k_{th}$ blocks;
                    \STATE Decouple the visual features $c_1^k$ with the shape of $[c_k,h_k,w_k]$ via Eq.~\ref{eq 2};
                    \STATE Decouple the non-visual features $c_2^k$ with the shape of $[c_k,h_k,w_k]$ via Eq.~\ref{eq 3};
                    \STATE Calculate visual discriminative features $\hat{f_i}$ via Eq.~\ref{eq 4};
                \ENDFOR
            \ENDFOR

            \STATE Calculate the  cross entropy loss $L_{cls}$ for $\hat{B}$;
            \STATE Back-pass and update $\theta$ and $\varphi$;
        \ENDWHILE
	\end{algorithmic} 
\end{algorithm}

\subsection{Training Design}

\textbf{Attack methods:} The adversarial examples are generated using state-of-the-art adversarial attack algorithms, including projected gradient descent (PGD) attack ($L_1$, $L_2$, $L_\infty$)~\cite{Madry2018Improving}, Decoupling Direction and Norm (DDN) attack ($L_2$)~\cite{rony2019decoupling},  Carlini $\&$ Wagner (C\&W) attack ($L_2$)~\cite{carlini2017towards}, AutoAttack (AA) attack ($L_\infty$)~\cite{zimmermann2019comment}, Auto-PGD (APGD) attack~\cite{zimmermann2019comment} ($L_2, L_\infty$), Fast adaptive boundary attack (FAB) attack ($L_2, L_\infty$)~\cite{croce2020minimally}, Square attack ($L_2, L_\infty$)~\cite{andriushchenko2020square}, Pixel attack ($L_0$)~\cite{pomponi2022pixle}, SparseFool attack ($L_0$)~\cite{modas2019sparsefool}, OnePixel attack ($L_0$)~\cite{su2019one},   expectation over transformation PGD (EOTPGD) attack ($L_\infty$) ~\cite{zimmermann2019comment} and Spatially-constrained (STA) attack ~\cite{xiao2018spatially}. 
These attacks are implemented by the advertorch library~\footnotemark~\cite{ding2019advertorch} and torch attacks library~\footnotemark~\cite{kim2020torchattacks}.

\footnotetext[1]{\href{https://github.com/BorealisAI/advertorch}{https://github.com/BorealisAI/advertorch}}
\footnotetext[2]{\href{https://github.com/Harry24k/adversarial-attacks-pytorch}{https://github.com/Harry24k/adversarial-attacks-pytorch}}

\begin{figure}[t]
\begin{center}
\includegraphics[width=0.75\linewidth]{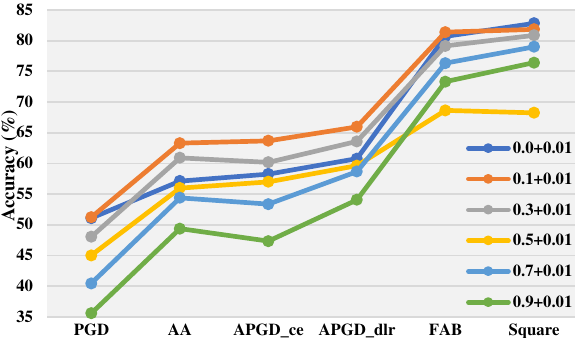} 
\caption{Parameter Analysis. The figure shows the classification accuracy rates (percentage) of different mask ratio defenses against different attacks (higher is better) on the CIFAR-10 dataset.}
\label{fig: parameter analysis}
\end{center}  
\end{figure}

\textbf{Implementation details:} We input the images of the CIFAR-10 dataset and CIFAR-100 dataset with $32 \times 32$, and the images of the Tiny-imagenet dataset with $64 \times 64$. Our proposed method is based on ResNet18. 
For the architecture design of Feature Decoupled Net, which is a simple three-layer network, we refer to the work of~\cite{Kim_2023_CVPR}.
We use an SGD optimizer to train the framework. The batch size of our experiments is 100. We follow the standard AT setting~\cite{Madry2018Improving} for the learning rate. After training the model, we add Decoupled Visual Feature Masking blocks in the individual layers and then fine-tune for 40 epochs (converges after about 10 epochs). All models are tested and trained on NVIDIA GeForce RTX 3090 24 GB GPU. We reported the accuracy (ACC) for classification results to evaluate the proposed method. Meanwhile, the parameters of our main attacks are as follows (the other attack parameter settings are given in supplementary materials):

\begin{itemize}

\item[$\bullet$] PGD ($L_\infty$): We set default perturbation budget = 8/255
for PGD. The default number of iterations is set to 40. The attack step size is set to 0.01. 

\item[$\bullet$] APGD ($L_\infty$): We set default perturbation budget = 8/255 for APGD. The default number of iterations is set to 10. we choose loss functions "ce" and "dlr".

\item[$\bullet$] Auto Attack ($L_\infty$): We set default perturbation budget = 8/255 for Auto Attack.

\item[$\bullet$] EOTPGD ($L_\infty$):  We set default perturbation budget = 8/255 for EOTPGD. The default number of iterations is set to 10. The attack step size is set to 2/255. 

\item[$\bullet$] FAB ($L_\infty$): We set default perturbation budget = 8/255 for FAB. The default number of iterations is set to 10.

\item[$\bullet$] DDN ($L_2$):  The default number of iterations is set to 40. The number of quantization levels is set to 256. The gamma is set to 0.05.

\item[$\bullet$] C\&W ($L_2$): The maximum number of iterations is set to 500. The confidence of the adversarial examples is set to 1. The Lr is set to 0.01.

\item[$\bullet$] Square ($L_\infty$): The maximum perturbation is set to 8/255. The max number of queries is set to 5000.

\item[$\bullet$] STA: The max iterations is set to 10. The search steps is set to 5.

\item[$\bullet$] Pixle, SparseFool and OnePixel follow the default Settings of the adversarial-attacks package.
\end{itemize}

\subsection{Parameter Analysis}~\label{Parameter Analysis}
The parameter settings used in this paper are as follows: We conduct a parameter study on CIFAR-10 data to evaluate the impact of $r_1, r_2$ and different blocks on adversarial robustness. To eliminate other confounding factors, we do not consider the impact of different blocks when evaluating the mask rate.
The parameters of the attack are given in the Experimental Settings section.

\begin{table*}
\centering
\caption{Classification accuracy (in percentage) for adversarial examples on CIFAR-10 (higher is better). For each attack, we show the most successful defense in bold and the second result in underline. If the DFM unit is inserted into the $i_{th}$ block, then we tick $B_i$.}
\begin{tabular}{cccc|ccccccccc} 
\hline
\multicolumn{4}{c|}{Block}  & \multicolumn{9}{c}{Classification accuracy for different  adversarial examples}     \\ 

\hline
B$_1$ & B$_2$ & B$_3$ & B$_4$           & None           & PGD            & DDN            & C\&W             & STA   & AA                & FAB             & Square      & \textcolor[rgb]{0,0,1}{Average}    \\ 
\hline
 \checkmark   & -   & -   &   -     & \underline{85.13}  & 52.38          & \underline{51.42}  & 59.48          & 21.12 & 62.05                             & \textbf{82.39} & 58.22     &59.05      \\
 -  & \checkmark     &-    &  -      & 83.51          & 52.49          & 51.19          & 58.96          & 25.82 & 65.28                          & 81.20           & 58.70      &59.64     \\
 -  & -   & \checkmark    &   -     & 83.39          & 52.53          & 50.26          & 61.17          & 28.68 & 67.75                        & 80.61           & 60.07     &  60.56    \\
-   &  -  &  -  & \checkmark        & \textbf{85.49} & 48.73          & 49.44          & 61.74          & 13.69 & 48.51                         & 75.98           & 47.35      & 53.87     \\
  \checkmark  & \checkmark    & -   & -    & 82.42          & \textbf{56.79}          & 47.76          & \textbf{63.01}          & 29.35 & \underline{72.92}                              & 81.39           &\textbf{ 66.89 }         & \underline{62.57}\\
  \checkmark  & -   &  -  & \checkmark    & 85.22          & 52.53          & \textbf{52.00} & 60.39          & 21.39 & 62.76                          & \underline{82.28}   & 58.37    &     59.37  \\
 -  & \checkmark    & -   & \checkmark     & 83.73          & 52.50          & 50.78          & 61.07          & 27.28 & 64.63                            & 81.18           & 58.60     & 59.97      \\

  \checkmark  & \checkmark    & -   & \checkmark    & 83.24          & \underline{56.61}          & 48.54          & \underline{62.38}          & \underline{28.27} & \textbf{73.03}                              & 82.21           & \underline{66.46}     & \textbf{62.59}    \\
\hline
\end{tabular}
\label{table: Ablation Study}
\end{table*}

\begin{table*}
\centering
\caption{Classification accuracy (in percentage) for adversarial examples on CIFAR-10 (higher is better). For each attack, we show the most successful defense in bold and the second result in underline.}
\scalebox{1.0}{
\begin{tabular}{ccccccccccccc} 
\hline
Methods  & Year                   & None           & PGD            & DDN            & C\&W             & STA            & AA             & APGD$_{ce}$           & APGD$_{dlr}$           & FAB            & Square         &\textcolor[rgb]{0,0,1}{Average}         \\ 
\hline
AT~\cite{Madry2018Improving}      & ICLR2018               & \textbf{85.64} & 47.93          & 1.11           & 9.61           & 0.04           & 48.22          & 48.22          & 50.28          & 49.75          & 53.28          & 39.41           \\
TRADES~\cite{zhang2019theoretically}    & ICML2019               & 80.88          & 51.69          & 0.11           & 47.70          & 1.17           & 49.40          & 51.57          & 49.40          & 50.72          & 53.19          & 43.58           \\
MART~\cite{Wang2020Improving}        & ICLR2020               & 82.38          & 52.50          & 1.47           & 28.26          & 1.70           & 50.26          & 52.43          & 50.26          & 50.55          & 53.41          & 42.32           \\
AWP~\cite{wu2020adversarial}         & NIPS2020               & 79.65          & 53.25          & 0.20           & 47.67          & 1.72           & 49.24          & 53.25          & 49.91          & 50.58          & 53.61          & 43.91           \\
Consist~\cite{tack2022consistency}  & AAAI2022               & 83.32          & 54.15          & 0.20           & 29.12          & 4.23           & 48.08          & 55.50          & 51.74          & 50.75          & 54.85          & 43.19           \\
FSR~\cite{Kim_2023_CVPR}     & CVPR2023               & 83.11          & 50.99          & 45.25          & \textbf{72.45} & 15.17          & 48.76          & 50.86          & 52.01          & 74.07          & 77.11          & 56.98           \\
AdaAD~\cite{Huang_2023_CVPR}    & CVPR2023               & \underline{}{85.07}  & 54.67          & 0.14           & 28.66          & 1.65           & 50.88          & 54.81          & 52.51          & 53.02          & 50.83          & 43.22           \\ 
\hline
AT       & \multirow{4}{*}{+DFM} & 83.24          & 56.61          & 48.54          & 62.38          & \underline{28.27}  & \underline{73.03}  & \textbf{70.96} & \underline{73.31}  & \textbf{82.21} & 66.46          & 64.50           \\
MART     &                        & 81.99          & \underline{57.39}  & 50.45          & 63.44          & \textbf{29.41} & 72.17          & 70.21          & 72.42          & \underline{80.84}  & \underline{81.28}  & \underline{65.96}   \\
TRADES   &                        & 81.97          & 53.41          & \textbf{52.48} & \underline{69.94}  & 22.51          & 59.76          & 59.99          & 59.83          & 77.54          & 80.37          & 61.78           \\
AdaAD    &                        & 83.94          & \textbf{59.53} & \underline{50.66}  & 68.37          & 25.85          & \textbf{74.46} & \textbf{72.92} & \textbf{74.47} & 74.77          & \textbf{83.61} & \textbf{66.86}  \\
\hline
\end{tabular}
}
\label{table: comparison result}
\end{table*}

\textbf{Influence of parameter mask ratio.} We conduct a parameter experiment on the CIFAR-10 dataset with different masking ratios. As shown in Figure~\ref{fig: parameter analysis}, we fix $r_1$ as $0.01$ and select the parameter $r_2$ from a set of \{ 0.0, 0.1, 0.2, 0.3, 0.4, 0.5, 0.6, 0.7, 0.8, 0.9 \}. We find that under natural samples, the accuracy continues to decline. At the same time, under different attack types, the accuracy of most adversarial samples shows a trend of first rising and then falling. 
Specifically, we find that the model performs best with parameter settings of 0.1 for $r_1$ and 0.01 for $r_2$.

\textbf{Influence of different blocks.} We explore the impact of inserting this unit at different blocks on the CIFAR-10 dataset. We divide ResNet18 into four blocks, and we use four DFM unit $\left\{F_{\varphi_1}, F_{\varphi_2}, F_{\varphi_3 }, F_{\varphi_4}\right\}$ to separate the visual features and non-visual features for different blocks. As shown in Table~\ref{table: Ablation Study}, supposing we insert our DFM unit $  F_{\varphi_k} $ in the $k_{th}$ block, we tick the B$_k$ in the table. The units at different blocks have inconsistent robustness to natural and adversarial samples. The combination of $\left\{F_{\varphi_1}, F_{\varphi_2}, F_{\varphi_4}\right\}$ can bring better adversarial robustness while natural accuracy will not have a large loss. Complete experimental results are available in Supp.

\subsection{Comparison Results}~\label{Comparison Results}
To verify the effectiveness of the proposed Decoupled Visual Feature Masking framework for improving adversarial robustness generalization, we evaluate our proposed method on CIFAR-10~\cite{krizhevsky2009learning}, CIFAR-100, and Tiny-imagenet~\cite{le2015tiny} datasets. The parameters of the attack are given in the section~\ref{Experiment setup}.

\textbf{Results on CIFAR-10 dataset.} We select some advanced methods, MART~\cite{Wang2020Improving}, TRADES~\cite{zhang2019theoretically}, AWP~\cite{wu2020adversarial}, Consist~\cite{tack2022consistency}, FSR~\cite{Kim_2023_CVPR}, AdaAD~\cite{Huang_2023_CVPR}, and standard AT~\cite{Madry2018Improving}, for comparison experiments. 
As shown in Table~\ref{table: comparison result},
we incorporate our method into existing AT methods. The plug-in can significantly increase the model’s generalization to adversarial samples while the performance drop is small.
Specifically, our method can achieve higher improvement after extension under PGD attack. Compared with the original method, the progress is 8.7 (AT), 4.89 (MART), 1.42 (TRADES), and 4.86 (AdaAD), respectively. At the same time, our method also has good defense performance compared with the plug-in method FSR. Only FSR, a plug-in method, has resistance regarding invisible attacks. And our method performs much better than the comparison methods.
It is worth noting that our method also performs well under AA attacks. Numerically, it even outperforms PGD attacks. This is because the attack intensity selected for AA attacks is less aggressive than those for PGD attacks. At the same time, our method also demonstrates good defensive capabilities against APGD.

\begin{table}
\centering
\caption{We train our approaches and comparison methods with PGD ($L_\infty$) of max perturbation = 8/255.  And we test the performance of our method and the comparison method under different perturbations (4/255,8/255,16/255).}
\scalebox{1.0}{
\begin{tblr}{
  cells = {c},
  hline{1,8} = {-}{0.08em},
  hline{2,6} = {-}{},
}
  Methods      & Dataset       & None & PGD   & DDN   & C\&W    \\
AT      & CIFAR-100     & 59.72  & 27.04 & 0.005 & 4.45  \\
MART & CIFAR-100 &41.52 &28.84 &0.25 &19.38\\
TRADES &CIFAR-100 &\textbf{55.94}&27.59 &0.11 &15.27\\
AT+DFM & CIFAR-100     & 52.25  & \textbf{29.81} & \textbf{34.62 }& \textbf{42.87 }\\
AT      & Tiny-imagenet & \textbf{31.15}  & 6.86  & 0.11  & 7.08  \\
AT+DFM & Tiny-imagenet & 29.56  & \textbf{8.41}  & \textbf{21.09} & \textbf{25.59} 
\end{tblr}
}
\label{table: ciar100}
\end{table}

\textbf{Results on CIFAR-100 and Tiny-imagenet datasets.} As shown in Table~\ref{table: ciar100}, we extend our method on standard AT; our method can increase the defense performance at PGD, while our method also greatly increases the defense performance against unseen attacks.
Specifically, the performance of my method is increased by 2.77\%  and 1.55\%, respectively, on PGD attacks. At the same time, the performance of DDN and C\&W attacks is improved by 20\%-30\%.

\begin{table*}
\centering
\caption{We train our approaches and comparison methods with PGD ($L_\infty$) of max perturbation = 8/255. We test the performance of our method and the comparison method under different perturbations (4/255,8/255,16/255).}
\begin{tabular}{cccc|cccc}
\hline
Methods  & 4/255                           & 8/255                           & 16/255                           & Methods    & 4/255                           & 8/255                           & 16/255                          \\ \hline
AT       & 69.80                           & 48.05                           & 14.39                            & TRADES     & 68.09                           & 51.74                           & \textbf{21.65}                           \\
AT+DFM   & \textbf{73.20} & \textbf{56.72} & \textbf{24.39}  & TRADES+DFM & \textbf{72.45} & \textbf{52.17} & 18.18 \\ \hline
MART     & 69.72                           & 52.47                           & 20.14                            & ADaAD      & 72.28                           & 54.72                           & 20.41                           \\
MART+DFM & \textbf{72.67} & \textbf{57.00} & \textbf{25.38} & ADaAD+DFM  & \textbf{75.42} & \textbf{59.15} & \textbf{26.70} \\ \hline
\end{tabular}
\label{table perturbation}
\end{table*}

\begin{table*}
\centering
\caption{We train our approaches  (LetNet + DFM) and comparison methods (LetNet) with PGD ($L_\infty$) of max perturbation = 0.4 in the MNIST dataset. We test the performance of our method and the comparison method under unseen attack.}
\begin{tabular}{lllllll} 
\hline
Architectures             & \multicolumn{1}{c}{Methods} & None           & PGD            & DDN             & C\&W              & STA             \\ 
\hline
\multirow{2}{*}{LetNet}   & AT                          & \textbf{98.81} & 95.35          & 59.14           & 97.22           & 9.86            \\
                          & AT+DFM                      & 98.57          & \textbf{95.55} & \textbf{ 67.54} & \textbf{97.81 } & \textbf{23.20}  \\ 
\hline
\multirow{4}{*}{ResNet18} & AT                          & \textbf{85.64} & 47.93          & 1.11            & 9.61            & 0.04            \\
                          & AT+DFM                      & 83.24          & 56.61          & 48.54           & 62.38           & \textbf{28.27}  \\
                          & AdaAD                       & 85.07          & 54.67          & 0.14            & 28.66           & 1.65            \\
                          & AdaAD+DFM                   & 83.94          & \textbf{59.53} & \textbf{50.66}  & ~\textbf{68.37} & 25.85           \\
\hline
\end{tabular}
\label{table MNIST}
\end{table*}


\textbf{Robust accuracy on different perturbation budgets.} 
We train our approaches and comparison methods with PGD ($L_\infty$) of max perturbation as 8/255. 
We test the performance of our method and the comparison method under different perturbation budgets (4/255,8/255,16/255).
As shown in Table~\ref{table perturbation}, with different perturbation budgets, our method also has better generalization.
Specifically, our method not only maintains the advantage of visible attacks but also has a better defense success rate in unseen perturbations.


\textbf{Defense against EOTPGD attack.}
We also discuss the defense performance of our method under EOTPGD attack, which is a more realistic and challenging attack scenario that considers the expectation over multiple transformations. Our defense method can significantly improve the defense performance under an EOTPGD attack. In general, the defense performance of EOTPGD with three layers of plugins can be increased by about 6\%. This demonstrates the effectiveness and robustness of our method against different types of adversarial attacks. Detailed experimental results can be found in the supplementary materials.




\begin{table*}
\centering
\caption{Ablation experiments of components. Classification accuracy (in percentage) for adversarial examples on CIFAR-10 (higher is better). For each attack, we show the most successful defense in bold. We treat our module as an image pre-processing operation instead of a plug-in unit. }
\begin{tabular}{ccccccc} 
\hline
Visual Feature & Non-visual Feature & None           & PGD            & CW             & STA            & \textcolor{blue}{Average}  \\ 
\hline
-              & -                  & \textbf{85.08} & 47.93          & 9.61           & 0.04           & 35.67                      \\
  \checkmark             & -                  & 84.30          & 51.08          & \textbf{62.30} & 19.37          & 54.26                      \\
   \checkmark            &        \checkmark            & 82.89          & \textbf{51.22} & 62.17          & \textbf{23.29} & \textbf{54.89}         \\ 
   \hline
\end{tabular}
\label{table:Ablation experiments 1}
\end{table*}


\textbf{Results on different architectures.}
In the MNIST dataset, we also add our proposed DFM unit to the LetNet network. As shown in Table~\ref{table MNIST}, Our approach can also improve the defense capability under visible attacks as well as within unseen attacks.

\textbf{Defense against EOTPGD attack.}
We also discuss the defense performance of our method under EOTPGD attack, which is a more realistic and challenging attack scenario that considers the expectation over multiple transformations. Our defense method can significantly improve the defense performance under EOTPGD attack. In general, the defense performance of EOTPGD with three layers of plugins can be increased by about 6\%. This demonstrates the effectiveness and robustness of our method against different types of adversarial attacks.

\textbf{Defense against different iteration steps.}
We train our approaches and comparison methods with PGD ($L_\infty$) of max perturbation as 8/255 and iteration steps as 20. 
And we test the performance of our method and the comparison method under iteration steps (20,40,100).
As shown in Table~\ref{table step}, with different perturbations, our method also has better generalization.
Specifically, our method not only maintains the advantage of visible attacks, but also has better defense success rate in unseen iteration steps.

\textbf{Results on Adaptive Attack.}
We discuss here the impact of adaptive attack on our approach. Since our method inserts the DFM unit on different blocks of the network. Compared with the BPDA~\cite{athalye2018obfuscated} attack method, $A^3$~\cite{Ye2022Practical} is more suitable for performance evaluation. It turns out that our method (AT+DFM) decreases from 50\% accuracy to 26\% accuracy on the CIFAR-10 dataset. However, the original method (AT) without plugin units shows almost no performance degradation. We think that a randomized defense approach will increase the expectation of robustness, while at the same time reducing the lower bound of robustness leads to poor results. Since our method focuses on having good generalization against various attacks, when it comes to defending against Adaptive Attacks, we only ensure that our method is competitive against other pre-processing defense methods.

\textbf{Analysis of model sizes.}
Here, we present the size of the original ResNet18 model, which is 44.70 MB, and the size of the ResNet18 model with the added DFM module, which is 48.39 MB. Our model size has increased by about 8\%, but it has significantly enhanced adversarial robustness by about 20\% on average.

\begin{table*}
\centering
\caption{We train our approaches and comparison methods with PGD ($L_\infty$) of max perturbation = 8/255. And we test the performance of our method and the comparison method under different iteration steps (20,40,100).}
\begin{tabular}{cccc|cccc} 
\hline
Methods  & 20             & 40             & 100            & Methods    & 20             & 40             & 100             \\ 
\hline
AT       & 48.26          & 47.96          & 47.90          & TRADES     & 51.82          & 51.74          & 51.67           \\
AT+DFM   & \textbf{58.40} & \textbf{56.44} & \textbf{56.34} & TRADES+DFM & \textbf{53.98} & \textbf{53.20} & \textbf{52.99}  \\ 
\hline
MART     & 52.62          & 52.47          & 52.42          & ADaAD      & 54.80          & 54.71          & 54.60           \\
MART+DFM & \textbf{58.20} & \textbf{56.86} & \textbf{56.76} & ADaAD+DFM  & \textbf{60.42} & \textbf{59.64} & \textbf{56.76}  \\
\hline
\end{tabular}
\label{table step}
\end{table*}

\section{Algorithm Analysis}~\label{Discussion}

\subsection{Ablation Study}~\label{Ablation Study}
The proposed Decoupled Visual Feature Masking framework comprises two design blocks: Decoupled Visual Masking and different depth-level features. To reveal the contribution of each module to performance improvement, we conduct a comprehensive ablation study and analyze them on the CIFAR-10 dataset, as shown in Table~\ref{table:Ablation experiments 1}. 
We utilize the AT as the baseline method for a fair comparison.
When only visual discriminative features are used, the accuracy under PGD attack is improved from 47.93\% to 51.08\%, and the defense performance against other invisible attacks is significantly improved. This is due to the fact that the decoupled visual representation helps separate the noise to increase the adversarial robustness.
When the network uses visual discriminative and non-visual features, the ACC under PGD attack improves by about 0.14\% compared to using only visual discriminative features; meanwhile, the defense performance against STA attack improves from 19.37\% to 23.29\%. Because our strategy not only disrupts adversarial noise but also ensures the integrality of discriminative information.
The ACC under PGD attack improves from 51.22\% to 56.47\% when different depth features are used, while the performance of other unseen attack defense methods also improves considerably: CW (3.01\%), STA (7.16\%).

\textbf{Ablation experiments of masking strategies.} To optimize the masking process, we proposed three different baselines, which mask different frequency and robustness regions. Specifically, the results are shown in Table~\ref{table:Ablation experiments 2} and  our three baselines are as follows:

\textit{S1:} Randomly mask the image with a masking ratio of 1\%. This method can disrupt some adversarial information and improve the model’s robustness. However, more information is needed, and the natural accuracy decreases greatly.

\textit{S2:} Separate the image into high-frequency and low-frequency regions and mask them with different ratios. This method can mask more high-frequency areas and less low-frequency regions according to the attack intensity of different frequencies. However, the partition is not optimal, and there is room for improvement.

\textit{S3:} Based on the second method, use the feature decouple network $F_\varphi$ to adaptively divide the feature into visual discriminative features and non-visual features and mask them with different ratios, respectively. The proposed visual masking strategy can help effectively improve the diversity of extracted robust features. Moreover, the following non-visual feature masking and fusion strategy not only disrupts adversarial noise but also ensures the integrality of discriminative information.

\begin{table}[t]
\centering
\caption{Ablation experiments of masking strategies. Classification accuracy (in percentage) for adversarial examples on CIFAR-10 (higher is better). For each attack, we show the most successful defense in bold.}~\label{table:Ablation experiments 2}
\scalebox{1}{
\begin{tblr}{
  hline{1,5} = {-}{0.08em},
  hline{2} = {-}{},
}
Methods  & None  & PGD   & DDN   & C\&W    & STA   \\
AT + S1 & 77.75 & 45.53 & 50.03 & \textbf{63.50} & \textbf{26.62} \\
AT + S2 & 82.26 & 50.06 & \textbf{51.63} & 61.76 & 22.28 \\
AT + S3  & \textbf{82.98} & \textbf{51.22} & 51.25 & 62.17& 23.29
\end{tblr}
}

\end{table}

\begin{table*}[t]
\centering
\caption{The average diversity of feature distribution between each image and the original image after multiple augmentations.
$\downarrow$ stands for "lower value is the larger the diversity of feature distribution between the original image and the transformed image," and $\uparrow$ stands for "higher value is the larger the diversity of feature distribution between the original image and the transformed image." }
\scalebox{1.0}{
\begin{tblr}{
  cells = {c},
  cell{2}{1} = {r=3}{},
  cell{5}{1} = {r=3}{},
  cell{8}{1} = {r=3}{},
  hline{1-2,5,8,11} = {-}{},
}
Methods & Scope         & PSNR$\downarrow$ & SSIM$\downarrow$ & MSE$\uparrow$ & Lipis$\uparrow$ & Std$\uparrow$ & KL dis$\uparrow$ & MA dis$\uparrow$ \\
Hue     & {[}-0.1, 0.1] & -                & 0.986            & 0.001         & 0.046           & 0.040         & 0.002            & 2.476            \\
        & {[}-0.2, 0.2] & -                & 0.961            & 0.004         & 0.105           & 0.071         & 0.004            & 7.502            \\
        & {[}-0.3, 0.3] & -                & 0.937            & 0.007         & 0.151           & 0.079         & 0.005            & 14.540           \\
Brightness     & {[}0.8, 1.2]  & 28.85            & 0.98             & 0.004         & 0.006           & 0.05          & 0.002            & 7.68             \\
        & {[}0.6, 1.4]  & 23.01            & 0.95             & 0.013         & 0.021           & 0.108         & 0.006            & 28.08            \\
        & {[}0.4, 1.6]  & 19.52            & 0.88             & 0.028         & 0.044           & 0.180         & 0.013            & 60.17            \\
contrast     & {[}0.8, 1.2]  & 36.73            & 0.99             & 0.001         & 0.004           & 0.036         & 0.002            & 1.27             \\
        & {[}0.6, 1.4]  & 30.78            & 0.99             & 0.002         & 0.014           & 0.080         & 0.004            & 4.85             \\
        & {[}0.4, 1.6]  & 27.57            & 0.94             & 0.005         & 0.032           & 0.137         & 0.008            & 10.40            
\end{tblr}
}
\label{table: data raodong}

\end{table*}

\subsection{Visualization Analysis}~\label{Visualization1}

\begin{figure*}
\begin{center}
\includegraphics[width=0.49\linewidth]{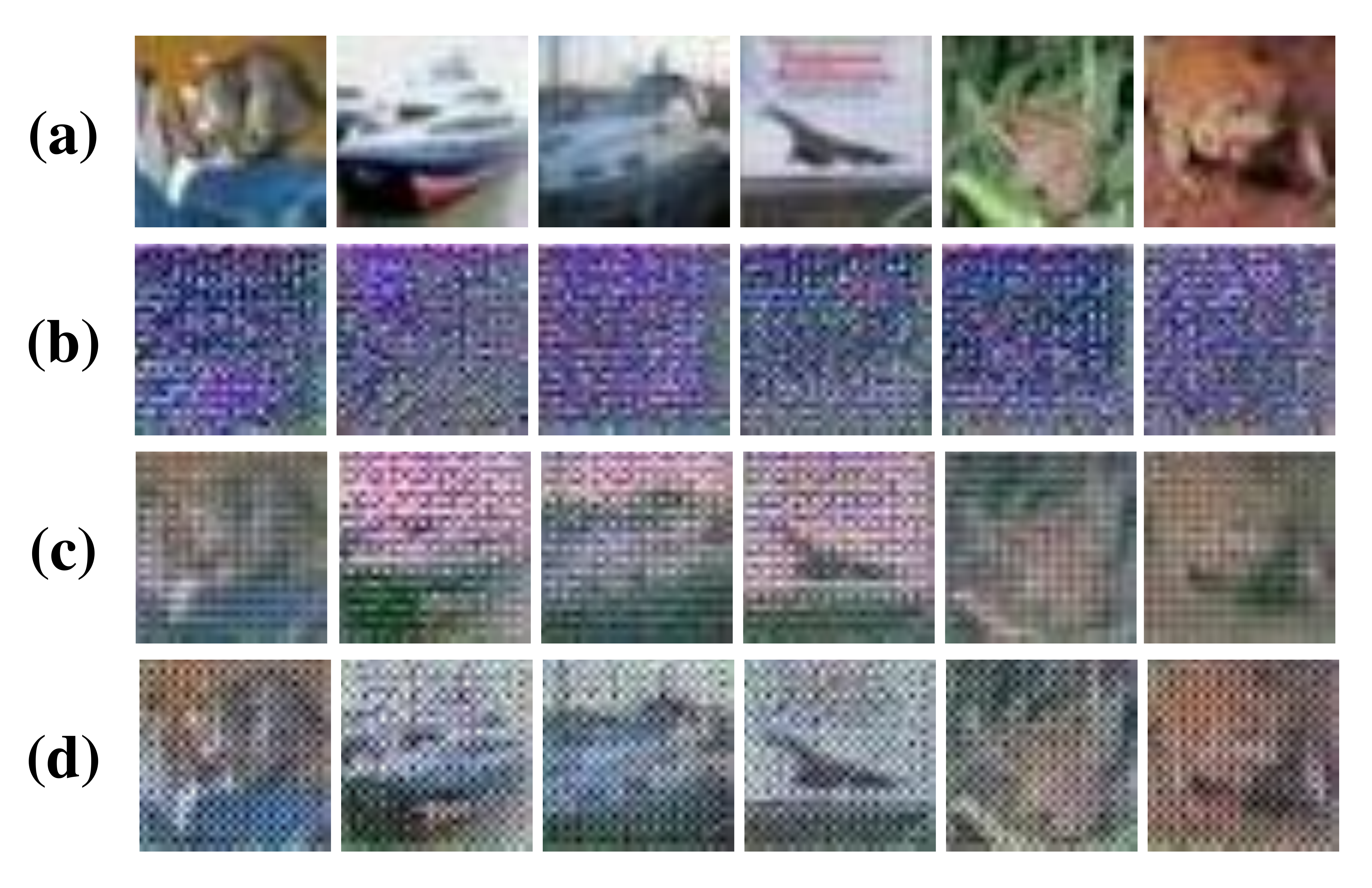} 
\includegraphics[width=0.49\linewidth]{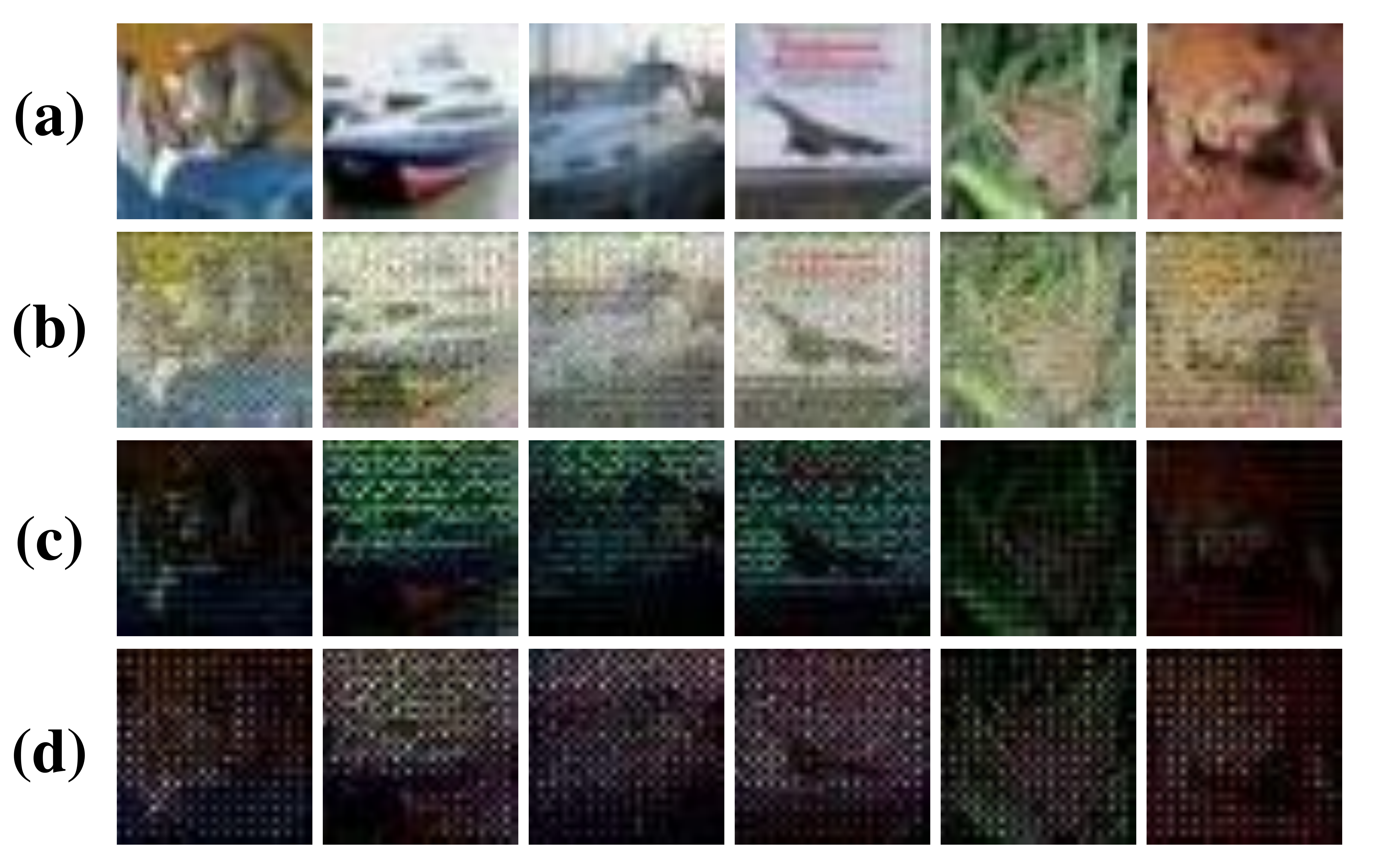} 
\caption{
\textbf{Left:}
(a) is the original image,
(b) is the reconstruction result of $r_1$ branch with $r_1 = 0.01$ and $r_2 = 0.0$, (c) is the reconstruction result of $r_1$ branch with $r_1 = 0.01$ and $r_2 = 0.5$, and (d) is the reconstruction result of $r_1$ branch with $r_1 = 0.01$ and $r_2 = 0.9$.
\textbf{Right:}
(a) is the original image,
(b) is the reconstruction result of $r_2$ branch with $r_1 = 0.01$ and $r_2 = 0.0$, (c) is the reconstruction result of $r_2$ branch with $r_1 = 0.01$ and $r_2 = 0.5$, and (d) is the reconstruction result of $r_2$ branch with $r_1 = 0.01$ and $r_2 = 0.9$.
}
\label{fig: discussion 2}
\end{center}  
\end{figure*}

To facilitate visualization, we treat our module as a pre-processing operation instead of plug-in units in visualization experiments.
We fix $r_1$ as $0.01$, select the parameter $r_2$ from a set of $\left\{0.0,0.5,0.9\right\}$, and observe the output results of the feature separation network $F_\varphi$. As shown in Fig~\ref{fig: discussion 2}, as $r_2$ increases, the visual discriminative information of $c_2$ gradually decreases. And the visual discriminative information of $c_1$ gradually increases.  This indicates that $F_\varphi$ can adjust the separation strategy according to different masking ratios and retain as much visual discriminative information as possible.
So, decoupling the two features does not need a specific loss function, which only depends on the masking strategies.
We just need to adjust the masking ratios $r_1$ and $r_2$ for visual discriminative features and non-visual features. 

\subsection{More Algorithm Analysis}\label{Visualization2}

\begin{figure}
\begin{center}
\includegraphics[width=\linewidth]{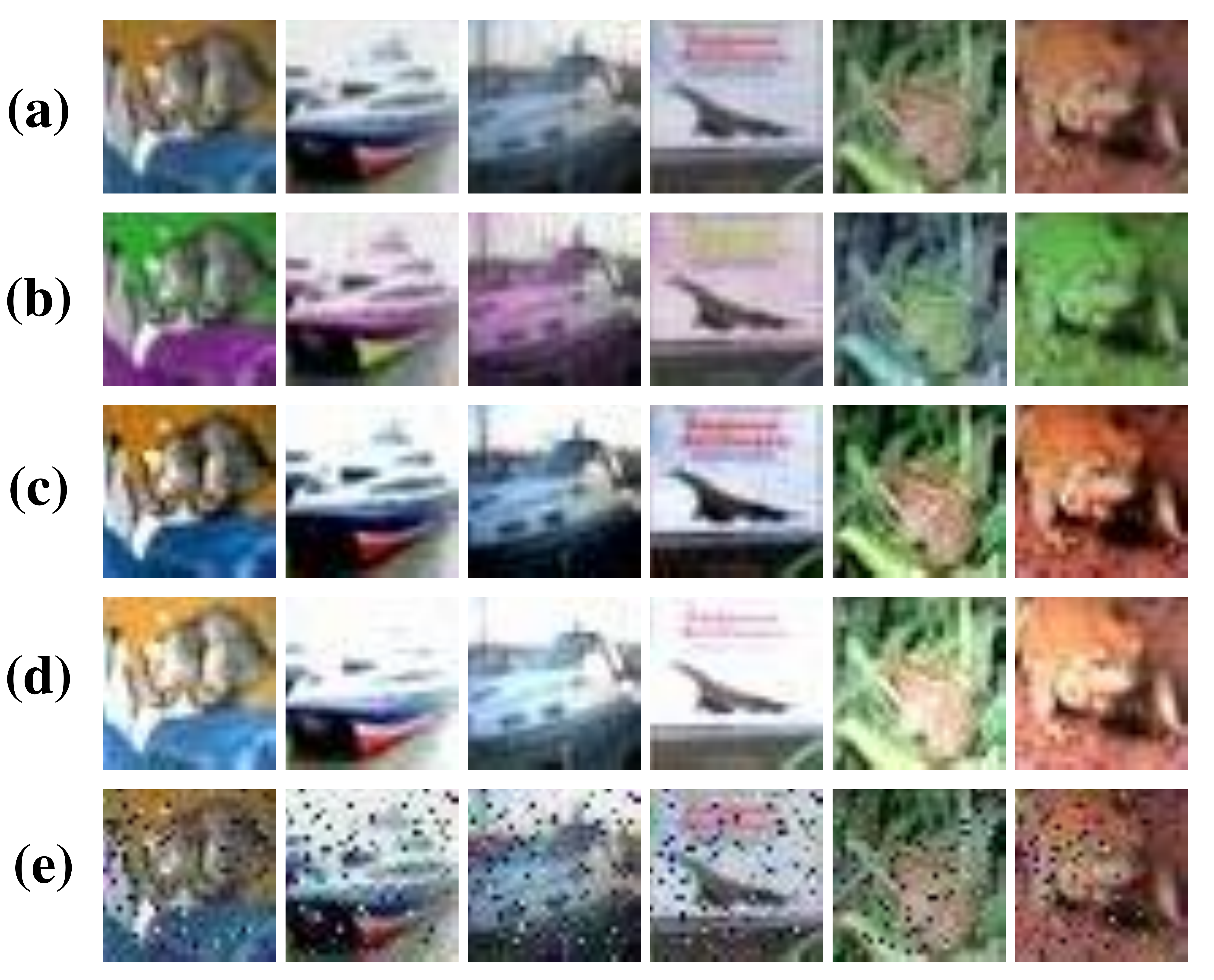} 
\caption{(a) is the original image, (b) is the image after color transformation, (c) is the image after contrast transformation, (d) is the image after brightness transformation, and (e) is our proposed method (for better visualization, we take image preprocessing methods as an example).}
\label{fig: data aug}
\end{center}  
\end{figure}

To explore the impact of data augmentation methods on adversarial robustness, we analyze three standard data augmentation methods: image hue change, image brightness change, and image contrast change. 
In Figure~\ref{fig: data aug}, we show the enhanced pictures of three standard data augmentation methods and the enhanced pictures of our proposed DFM unit.
\textbf{ For better visualization, we consider our method also as an image preprocessing method as a simplified version.}
Table~\ref{table analysis} shows the improvement of adversarial robustness after applying different data augmentation methods in the testing phase. Three standard data augmentation methods improve adversarial robustness and significantly affect invisible attacks. Specifically, increasing image distortion can substantially enhance the model’s adversarial robustness for STA, C\&W, and AA attacks. 
The data variance increases significantly as the masking rate increases, indicating that the data distribution becomes more dispersed. Finding a universal attack noise that worked for all samples was difficult, which is conducive to improving the model’s adversarial robustness. 
Meanwhile, for a PGD attack, choosing a moderate image distortion can better boost the model’s defense ability against PGD. This is because when the distortion rate is too high, the data variance is too large, which hinders the model from fitting the actual data distribution. This will lower the model’s accuracy on natural and adversarial samples. Therefore, increasing data variance can improve adversarial robustness, but choosing an appropriate distortion rate is still necessary.

\begin{table}[t]
\centering
\caption{After adding image preprocessing, the adversarial robustness is improved (\%) on CIFAR-10 (higher is better). For each attack, we show the most successful defense in bold.}
\scalebox{1.0}{
\begin{tblr}{
  cells = {c},
  cell{2}{1} = {r=3}{},
  cell{5}{1} = {r=3}{},
  cell{8}{1} = {r=3}{},
  hline{1-2,5,8,11} = {-}{},
}
Methods & Scope         & PGD           & DDN            & C\&W              & STA            & AA             \\
Hue     & {[}-0.1, 0.1] & 1.58          & 50.81          & 56.09           & 15.26          & 7.61           \\
        & {[}-0.2, 0.2] & 2.14          & \textbf{52.56} & 50.62           & 19.88          & 12.73          \\
        & {[}-0.3, 0.3] & 2.05          & 44.37          & 51.44           & 21.14          & 13.38          \\
Brightness     & {[}0.8, 1.2]  & 2.49          & 50.52          & 50.53           & 20.00          & 7.52           \\
        & {[}0.6, 1.4]  & \textbf{3.26} & 50.73          & 53.85           & 26.11          & \textbf{13.80} \\
        & {[}0.4, 1.6]  & 1.89          & 51.63          & 56.32           & \textbf{31.92} & 12.02          \\
contrast     & {[}0.8, 1.2]  & 1.79          & 51.46          & 51.02           & 19.16          & 5.31           \\
        & {[}0.6, 1.4]  & 2.31          & 51.80          & 52.18           & 24.73          & 10.84          \\
        & {[}0.4, 1.6]  & 0.16          & 50.45          & \textbf{ 57.38} & 27.61          & 12.64          
\end{tblr}
}
\label{table analysis}
\end{table}

\subsection{Future Directions}
In this section, we discuss several possible future research directions based in adversarial preprocessing defense to inspire research in the community.

One research direction is to explore effective preprocessing defense methods under expected adversarial attacks. Some studies have shown that most random defense methods are easily broken under expectation attacks. At the same time, these methods lack sufficient randomness. Maintaining effective adversarial defense under expectation attacks is a crucial point worth studying.

Another research direction is to explore the trade-off of preprocessing defense methods on natural samples and adversarial robustness. Random perturbation can improve the ability of adversarial defense, but it also increases the difficulty of fitting the model to the data. Therefore, it is worth considering how to design preprocessing methods that are friendly to natural samples and effective for adversarial samples.

The third research direction is to explore the trade-off between the efficiency and effectiveness of preprocessing defense methods. Some preprocessing algorithms, such as denoising or super-resolution, may significantly increase computational resources, thus limiting their usability in real-world scenarios. Therefore, it is desirable to design preprocessing defense methods that achieve both high robustness and low overhead.

\section{Conclusion}

This paper highlights two novel properties of robust features from the feature distribution perspective: diversity and discriminability. 
Based on this discovery,
we propose an effective defense based on decoupled visual representation masking. The designed Decoupled Visual Feature Masking (DFM) block can adaptively disentangle visual discriminative features and non-visual features with diverse mask strategies, while the suitable discarding information can disrupt adversarial noise to improve robustness. Our work provides a generic and easy-to-plugin block unit for any former adversarial training algorithm to achieve better protection integrally.
Experimental results show that the method has good defense performance under PGD attack and good generalization ability under invisible attack.

%

\bibliographystyle{IEEEtran}
\bibliography{egbib}

\end{document}